\pdfoutput=1

\documentclass[11pt]{article}

\usepackage{acl}

\usepackage{times}
\usepackage{latexsym}

\usepackage{amsmath,amssymb,amsfonts}
\usepackage{algorithmic}
\usepackage{graphicx}
\usepackage{multirow}
\usepackage{multicol}
\usepackage{xcolor}
\graphicspath{{figures/}}

\usepackage[T1]{fontenc}

\usepackage[utf8]{inputenc}

\usepackage{microtype}

\title{Utterance Classification with Logical Neural Network: 
\\ Explainable AI for Mental Disorder Diagnosis}

\author{
Yeldar Toleubay \\ IBM Research, Japan \\{yeldar.toleubay@gmail.com  } \\ \And
Don Joven Agravante \\IBM Research, Japan \\
{   don.joven.r.agravante@ibm.com}\\ \And
Daiki Kimura\\IBM Research, Japan \\{daiki@jp.ibm.com }  
\AND
Baihan Lin \\
Columbia University, USA \\
{baihan.lin@columbia.edu  } \\ \And
Djallel Bouneffouf\\
IBM Research, USA \\
{djallel.bouneffouf@ibm.com  } \\ \And
Michiaki Tatsubori \\
IBM Research, Japan \\
{mich@jp.ibm.com } }
% ... 

\begin{document}
\maketitle
\begin{abstract}
In response to the global challenge of mental health problems, we proposes a Logical Neural Network (LNN) based Neuro-Symbolic AI method for the diagnosis of mental disorders. Due to the lack of effective therapy coverage for mental disorders, there is a need for an AI solution that can assist therapists with the diagnosis. However, current Neural Network models lack explainability and may not be trusted by therapists. The LNN is a Recurrent Neural Network architecture that combines the learning capabilities of neural networks with the reasoning capabilities of classical logic-based AI. The proposed system uses input predicates from clinical interviews to output a mental disorder class, and different predicate pruning techniques are used to achieve scalability and higher scores. In addition, we provide an insight extraction method to aid therapists with their diagnosis. The proposed system addresses the lack of explainability of current Neural Network models and provides a more trustworthy solution for mental disorder diagnosis.
\end{abstract}

\section{Introduction}

A mental disorder is a significant deterioration of human thinking, emotional control, or behavior, which is diagnosed clinically and can affect key areas of life. Due to the COVID-19 pandemic, the number of people who suffer from anxiety and depressive illnesses greatly increased in 2020. Initial projections indicate a 26\% and 28\% increase in anxiety and major depressive disorders respectively during the first year of the pandemic \cite{who}. Moreover, every year, 703 000 people commit suicide, with many more attempting to do so.  Although people of all ages commit suicides, it is alarming that in 2019 suicide was one of the leading causes of death among young people worldwide \cite{Suicide}. Furthermore, around 24 million people, or 1 in 300 persons (0.32\%), globally suffer from schizophrenia. Although it is not common as other mental disorders, schizophrenia produces psychosis, is associated with significant disability, and may have an impact on all aspects of life, including personal, family, social, educational, and occupational functioning \cite{Schizophrenia}. 

Diagnosis of mental disorders is accomplished through a clinical interview, where a therapist evaluates the mental health of the patient and identifies possible disorders based on symptoms. However, although many mental health issues may be properly treated at low cost, there is still a wide gap between those who need care and those who have access to it. Despite the progress in some countries, there is still a severe lack of effective therapy coverage. Therefore, there is a need for an AI solution that can assist therapists with a diagnosis of mental disorders. 

Although current Neural Network (NN) models are powerful and can operate in a wide range of tasks, their effectiveness in mental disorder classification is questionable due to their black-box nature. In this regard, the model explainability is a vital property, which is required to make a diagnosis of mental disorders. While  Neural Network models can achieve high scores, therapists may be hesitant to trust such tools and accept classification results if proper explanations are not provided. Because of NN nature, it is impossible to tell whether their predictions are the result of robust features or some spurious clues \cite{Ribeiro}.  There are attempts to provide interpretable insights in mental disorder diagnosis, such as using topic modeling to extract concepts \cite{lin2022neural} or inferring psychological properties such as working alliance \cite{lin2022deep}. Although such approaches can enable explainable AI systems for passive assistance \cite{lin2023helping,lin2022voice2alliance} or interventional recommendations \cite{lin2022supervisor,lin2023psychotherapy} to the therapists, applying these insights directly to the classification problem yields suboptimal performance \cite{lin2022working}. Furthermore, despite being able to provide global explanations for the prediction \cite{Mowery2017UnderstandingDS},  traditional ML models lack scalability and they are not generalizable for broader tasks.

In this regard, Logical Neural Network \cite{LNN} might be a good solution to the problem. It is a Neuro-Symbolic AI method (NSAI),  that combines the learning capabilities of neural networks with the reasoning capabilities of classical logic-based AI. The LNN is a Recurrent neural network architecture in which neurons represent a precisely defined notion of weighted real-valued logic. It has a 1-to-1 relationship to a system of logical formulae. The main problem related to this approach is that it has not been implemented for the supervised learning utterance classification task. Therefore, this work proposes an LNN-based explainable NSAI utterance classification method for mental disorder diagnosis. The model was trained with different predicate pruning techniques to achieve scalability and higher scores. The advantages of the proposed system can be summarized via the following points: 
\begin{itemize}
    \item We propose design of the supervised NSAI method for utterance classification task, where input to the model is predicates from clinical interviews and output is a mental disorder class. After the training the system outputs weighted logical rule to make classifications.
    \item We propose a predicate pruning methods to improve scalability and generalizability of the model.
    \item We propose an insight extraction methods which can aid therapists with their mental disorder diagnosis.
    
\end{itemize}

\begin{figure*}[h]
    \centering
    \includegraphics[width=0.85 \textwidth]{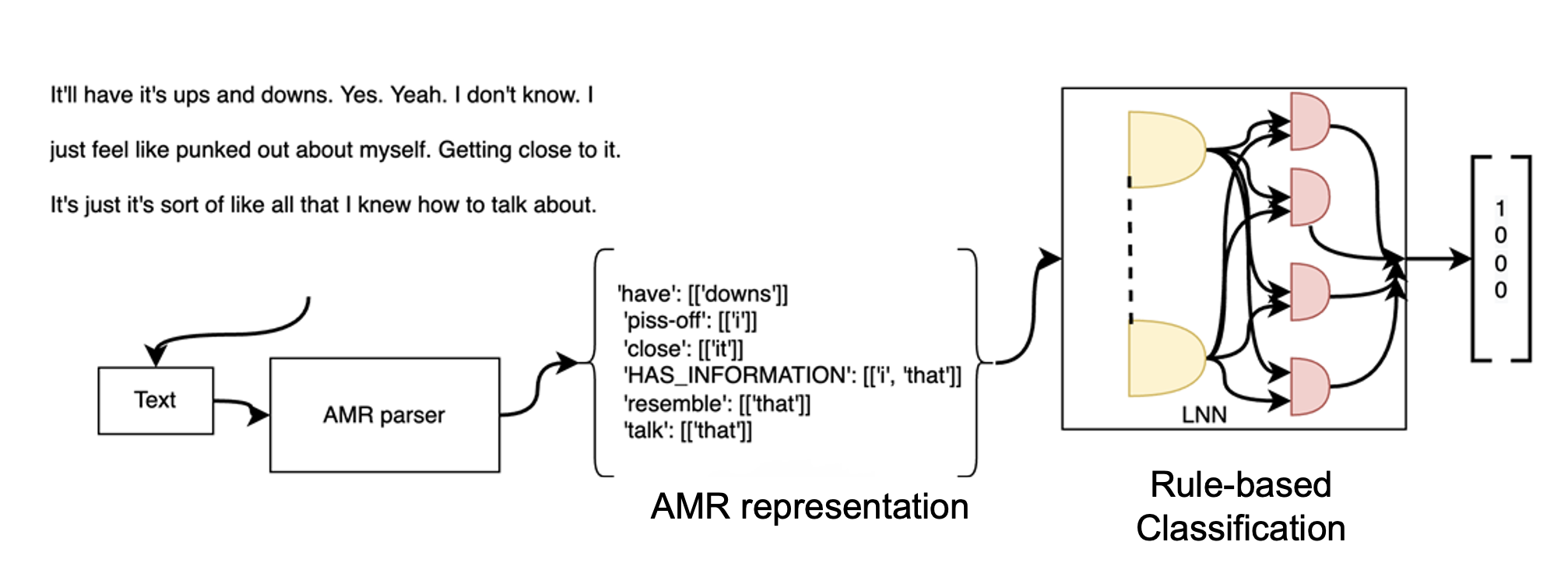}
    \caption{The overview of the proposed system.  }
    \label{fig:system}
\end{figure*}

1)  This paper is organized as follows:  Section II details the proposed system, Section III contains experiment results , Section IV provides discussions and future work, and the paper ends with a Conclusion.



\section{Supervised learning with LNN}

Although NSAI supports data driven training of the network, it encodes knowledge into logic rules with predicates as inputs, where predicates represents a property or a relation. Therefore, NSAI method requires special preprocessing of the dataset to generate predicates and data samples for training and testing purposes. The proposed system consists of two parts Abstract Meaning Representation (AMR) \cite{AMR} semantic parser and LNN. Fig. \ref{fig:system} shows overall pipeline of the system, first part containing AMR parser is used to convert raw text into classifier input data, and second part is an LNN model which performs rule-based classification. 

\subsection{Dataset preperation and preprocessing}

\begin{figure}[h]
    \centering
    \includegraphics[width=0.3\textwidth]{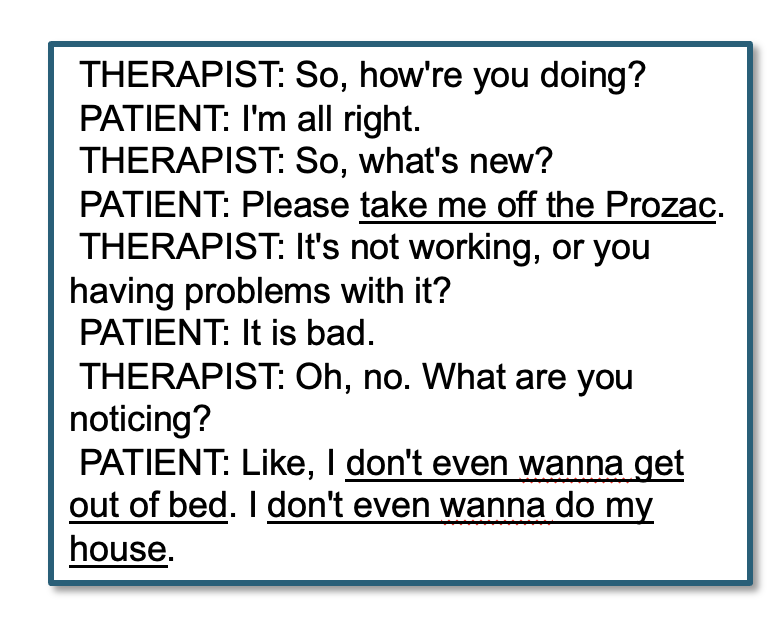}
    \caption[Examples of a dataset transcript.]{Examples of a dataset transcript.}
    \label{fig:transcript}
\end{figure}

Counseling and Psychotherapy Transcripts \cite{alexanderstreet} is a unique and fully anonymized online series of clinical interviews that allows students and researchers to dive deeply into the patient-therapist relationship and track the progress and setback of patients over multiple therapy sessions. These materials bring the mental disorder diagnosis process to life and provide unprecedented levels of access to the widest possible range of clients. Therefore, transcripts of 4 types of mental disorders, which are anxiety, depression, suicidal thoughts, and schizophrenia,  from this dataset are used in our training and evaluation of the model. Table \ref{table:dataset} shows the details of the dataset; in our simulations, only 12 sessions of clinical interviews have been used due to the computational constraints of semantic parsing. An example from the transcript has shown in Fig \ref{fig:transcript}. In our experiments, a transcript is a full clinical interview between a patient and a therapist, while an utterance represents a full response of the patient to a specific question from the therapist.

\begin{table}[h]
\centering
\begin{tabular}{|l|l|l|}
\hline
\textbf{Class}         & \textbf{\begin{tabular}[c]{@{}l@{}}Number of \\ total sessions\end{tabular}} & \textbf{\begin{tabular}[c]{@{}l@{}}Number of \\ used session\end{tabular}} \\ \hline
\textit{Anxiety}       & 498                                                                          & 12                                                                         \\ \hline
\textit{Depression}    & 377                                                                          & 12                                                                         \\ \hline
\textit{Suicidal}      & 12                                                                           & 12                                                                         \\ \hline
\textit{Schizophrenia} & 71                                                                           & 12                                                                         \\ \hline
\end{tabular}
\caption{Details of the dataset.}
\label{table:dataset}

\end{table}

As mentioned before, LNN requires a special data structure to function. AMR parser is used for generation of predicates by extracting the semantics of the utterance and converting  semantics into a graph, where nodes (keys) represent concepts and edges (values) represent relations to concepts. Example of AMR Representation is shown in Fig \ref{fig:system}. AMR Representation keys and values are combined to generate predicates as shown in Table \ref{table:inputsoutputs}. 

\begin{table*}[h]
\centering
\begin{tabular}{|c|lll|l|}
\hline
\textbf{Input}                & \multicolumn{2}{c|}{\textbf{Predicates}}                                                & \multicolumn{1}{c|}{}                                                            & \textbf{Output}                                    \\ \hline
                          & \multicolumn{3}{c|}{AMR Representation}                                                                                                                           &                                                \\ \hline
\multicolumn{1}{|l|}{}    & \multicolumn{1}{l|}{Keys}   & \multicolumn{1}{l|}{Values}             & \multicolumn{1}{c|}{\begin{tabular}[c]{@{}c@{}}Output of \\ Parser\end{tabular}} & Class                                          \\ \hline
\multirow{12}{*}{sample0} & \multicolumn{1}{l|}{HAS\_POSSESSION} & \multicolumn{1}{l|}{your medication}    & TRUE                                                                             & \multirow{12}{*}{Depression}                   \\ \cline{2-4}
                          & \multicolumn{1}{l|}{HAS\_POSSESSION} & \multicolumn{1}{l|}{any details}        & FALSE                                                                            &                                                \\ \cline{2-4}
                          & \multicolumn{1}{l|}{HAS\_POSSESSION} & \multicolumn{1}{l|}{downs}              & FALSE                                                                            &                                                \\ \cline{2-4}
                          & \multicolumn{1}{l|}{HAS\_POSSESSION} & \multicolumn{1}{l|}{just awkward thing} & FALSE                                                                            &                                                \\ \cline{2-4}
                          & \multicolumn{1}{l|}{have}            & \multicolumn{1}{l|}{your medications}   & FALSE                                                                            &                                                \\ \cline{2-4}
                          & \multicolumn{1}{l|}{have}            & \multicolumn{1}{l|}{any details}        & FALSE                                                                            &                                                \\ \cline{2-4}
                          & \multicolumn{1}{l|}{have}            & \multicolumn{1}{l|}{downs}              & TRUE                                                                             &                                                \\ \cline{2-4}
                          & \multicolumn{1}{l|}{have}            & \multicolumn{1}{l|}{just awkward thing} & FALSE                                                                            &                                                \\ \cline{2-4}
                          & \multicolumn{1}{l|}{talk}            & \multicolumn{1}{l|}{your medication}    & FALSE                                                                            &                                                \\ \cline{2-4}
                          & \multicolumn{1}{l|}{talk}            & \multicolumn{1}{l|}{any details}        & FALSE                                                                            &                                                \\ \cline{2-4}
                          & \multicolumn{1}{l|}{talk}            & \multicolumn{1}{l|}{downs}              & FALSE                                                                            &                                                \\ \cline{2-4}
                          & \multicolumn{1}{l|}{talk}            & \multicolumn{1}{l|}{just awkward thing} & FALSE                                                                            &                                                \\ \hline
\multirow{12}{*}{sample1} & \multicolumn{1}{l|}{HAS\_POSSESSION} & \multicolumn{1}{l|}{your medications}   & TRUE                                                                             & \multicolumn{1}{c|}{\multirow{12}{*}{Anxiety}} \\ \cline{2-4}
                          & \multicolumn{1}{l|}{HAS\_POSSESSION} & \multicolumn{1}{l|}{any details}        & TRUE                                                                             & \multicolumn{1}{c|}{}                          \\ \cline{2-4}
                          & \multicolumn{1}{l|}{HAS\_POSSESSION} & \multicolumn{1}{l|}{downs}              & FALSE                                                                            & \multicolumn{1}{c|}{}                          \\ \cline{2-4}
                          & \multicolumn{1}{l|}{HAS\_POSSESSION} & \multicolumn{1}{l|}{just awkward thing} & TRUE                                                                             & \multicolumn{1}{c|}{}                          \\ \cline{2-4}
                          & \multicolumn{1}{l|}{have}            & \multicolumn{1}{l|}{your medications}   & FALSE                                                                            & \multicolumn{1}{c|}{}                          \\ \cline{2-4}
                          & \multicolumn{1}{l|}{have}            & \multicolumn{1}{l|}{any details}        & FALSE                                                                            & \multicolumn{1}{c|}{}                          \\ \cline{2-4}
                          & \multicolumn{1}{l|}{have}            & \multicolumn{1}{l|}{downs}              & FALSE                                                                            & \multicolumn{1}{c|}{}                          \\ \cline{2-4}
                          & \multicolumn{1}{l|}{have}            & \multicolumn{1}{l|}{just awkward thing} & FALSE                                                                            & \multicolumn{1}{c|}{}                          \\ \cline{2-4}
                          & \multicolumn{1}{l|}{talk}            & \multicolumn{1}{l|}{your medications}   & FALSE                                                                            & \multicolumn{1}{c|}{}                          \\ \cline{2-4}
                          & \multicolumn{1}{l|}{talk}            & \multicolumn{1}{l|}{any details}        & TRUE                                                                             & \multicolumn{1}{c|}{}                          \\ \cline{2-4}
                          & \multicolumn{1}{l|}{talk}            & \multicolumn{1}{l|}{downs}              & FALSE                                                                            & \multicolumn{1}{c|}{}                          \\ \cline{2-4}
                          & \multicolumn{1}{l|}{talk}            & \multicolumn{1}{l|}{just awkward thing} & FALSE                                                                            & \multicolumn{1}{c|}{}                          \\ \hline
\end{tabular}

\caption{LNN for supervised learning inputs and outputs – predicates, data samples and class}
\label{table:inputsoutputs}
\end{table*}

Moreover, a training and a testing sample is input to the model and is obtained by using AMR parser over an utterance. Furthermore, a sample contains all predicates that has been mined from dataset and the corresponding output of parser as groundings. The values of the groundings are assigned according to the presence of the particular predicates in the parsed utterance, which means only predicates that results from that particular utterance assigned with \textit{TRUE} grounding for that particular sample. In this regard, certain combinations of predicates might repeat over multiple classes and the proposed design takes into account this issue.

\subsection{Proposed system details}

LNN is a core of the model, which has only few differences from regular neural network. The main difference of LNN is that its neural parameters  are limited such that the truth functions of the relevant logical gates govern the behavior of the neurons. Moreover, LNN neuron has more parameters compared to dense neuron, since it keeps both upper and lower bounds to the  corresponding subformula or predicate. 

\begin{figure}[]
    \centering
    \includegraphics[width=0.3\textwidth]{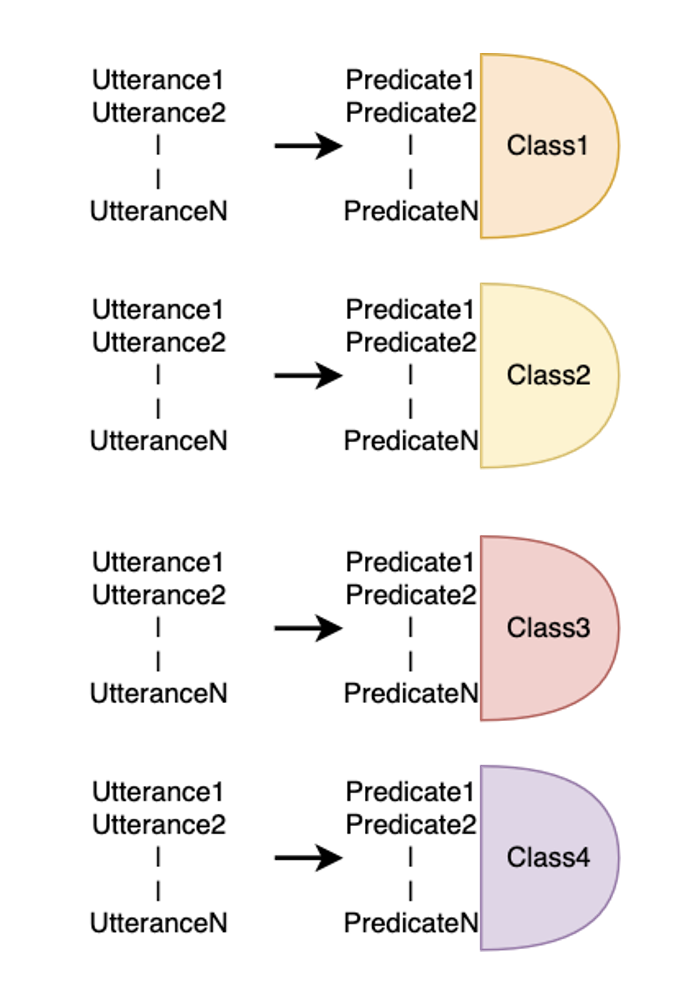}
    \caption[Proposed LNN architecture for mental disorder diagnosis.]{Proposed LNN architecture for mental disorder diagnosis.}
    \label{fig:LNN}
\end{figure}

The proposed LNN architecture has 4 \textit{AND} logic gates that act as binary classifiers for each mental disorder class. Predicates are inputs to the logic gates, while model is trained by samples generated from utterances. Those samples show truth values for formulae. After the training model outputs set of weight for each predicate and  outputs a tensor of lower and upper bounds as a score for a particular input.  In our experiments the each logic gate is evaluated as a binary classifier that classifies according to some threshold, thus the upper and lower bounds are averaged to obtain a single score. The score \textit{S} for each class is obtained via following equation:

\begin{equation}
    S=w_1\ P_1\ (x_1)+w_2\ P_2\ (x_2)+..+w_N\ P_N\ (x_N)
\end{equation}
 
where $P$ is a predicate, $w$ is a weight obtained from training and $x_i$ is a grounding for each predicate in a sample. 

The proposed system will be evaluated as a separate binary classification models for each gate by True Positive Rate (TPR)  and False  Positive Rate (FPR) metrics. The TPR indicates the proportion of all available positive samples that contain correct positive results. In contrast, FPR quantifies the proportion of available negative samples that contain incorrectly positive results. Moreover, the receiver operating characteristic (ROC)  curve is created by plotting the TPR against the FPR at various threshold values.

\begin{equation}
    TPR=\frac{True\ Positives}{True\ Positives\ +False\ Negatives}
\end{equation}

\begin{equation}
    FPR=\frac{False\ Positives}{True\ Negatives\ +False\ Positives}
\end{equation}

\subsection{Predicate pruning methods}

\begin{table}[h]
\centering
\begin{tabular}{|c|c|}
\hline
\textbf{\# of predicates} & \textbf{\begin{tabular}[c]{@{}c@{}}Training time\\ (s)\end{tabular}} \\ \hline
710                       & 4.49                                                                 \\ \hline
1415                      & 16.54                                                                \\ \hline
\end{tabular}

\caption{ Results of training time with different number of predicates for an LNN model with 2 Logic gates.}
\label{table:time}
\end{table}

\begin{table*}[h]
\centering
\begin{tabular}{|l|c|c|c|c|c|c|c|}
\hline
\multicolumn{1}{|c|}{\textbf{Class}} & \textbf{\begin{tabular}[c]{@{}c@{}}Original \\ predicates\end{tabular}} & \textbf{\begin{tabular}[c]{@{}c@{}}Similarity\\ pruning\end{tabular}} & \textbf{\begin{tabular}[c]{@{}c@{}}Exclusive\\ Pruning\end{tabular}} & \textbf{F=1} & \textbf{F=2} & \textbf{2\textless{}F\textless{}10} & \textbf{F\textgreater{}9} \\ \hline
\textit{Anxiety}                     & 5529                                                                    & 2773                                                                  & 245                                                                  & 3152         & 216          & 150                                 & 14                        \\ \hline
\textit{Depression}                  & 7227                                                                    & 3532                                                                  & 472                                                                  & 2174         & 454          & 133                                 & 12                        \\ \hline
\textit{Suicidal}                    & 6067                                                                    & 3213                                                                  & 230                                                                  & 2839         & 197          & 160                                 & 17                        \\ \hline
\textit{Schizophrenia}               & 3746                                                                    & 1914                                                                  & 96                                                                   & 1718         & 102          & 87                                  & 7                         \\ \hline
\end{tabular}

\caption{Number of original predicates and number of predicates after  similarity, exclusive and frequency pruning methods.}
\label{table:numberofpredicates}
\end{table*}

Predicates play a crucial role in LNN training and can greatly affect the accuracy of the model. Table \ref{table:numberofpredicates} shows that 48 transcripts result in more than 19000 predicates. However, according to Table \ref{table:time} a preliminary simulation results show that for a linear increase in number of predicates, LNN requires exponential increase in training time. Therefore, there is a need for predicate pruning methods, which will help to chose predicates that contribute the most towards the correct diagnosis. Thus, similarity, exclusivity and frequency based predicate pruning methods has been proposed to reduce number of predicates. 

\textit{Similarity pruning}. Simulations has shown that AMR Parser returns multiple variants of values per one key. Often, those values contain repeating phrases. Thus, it is possible to group all those lookalike predicates by taking a predicate that contains possible repetitions, e.g. instead of taking both “HAS\_POSSESSION\_my sister’s birthday” and “HAS\_POSSESSION\_sister’s birthday”, one can take only the first one.

\textit{Frequency pruning}. In traditional ML word count can show the importance of some features for a specific class. Using the same logic, it was assumed that predicates that are encountered frequently in sessions will have higher impact on model training. Thus, predicates has been analyzed in terms of repetitions across sessions and have been grouped according to the specified frequencies.
 
\textit{Exclusive pruning}. Since transcripts are  conversations between patients and therapists, there are many repeating predicates between classes. Thus, it was suggested that predicates belonging only to a class will avoid contradictions in the model as well as will have higher correlation to a specific class. Therefore, predicates repeating between classes predicates that are repeated only once have been removed.

\section{Experiment results}
In this section experiment results for predicate pruning and LNN model evaluation will be provided. Table \ref{table:numberofpredicates} shows number of predicates for a particular pruning method. Similarity pruning method prunes almost half of the original predicates. Furthermore, Exclusive and Frequency pruning methods have been applied on top of the similarity pruning method. Results for the Exclusive pruning shows that Depression class has twice of Anxiety and Suicidal predicates and 5-times  of Schizophrenia predicates. Moreover, results for the different frequencies show that majority of the predicates (43\%) repeat just once, while the higher frequency rates have fewer predicates. 

The LNN models have been trained using different pruning methods and have been compared with Deep Learning (DL) and LNN baselines. The number of predicates and training samples are shown in the Table \ref{table:auc}. The LNN models have been trained with supervised loss, which targets the labels with learning rate of 0.05, for 50 epochs. The main difference between LNN models is in the predicates. The details of each model are summarized below:

\begin{itemize}
    \item \textit{DL baseline}. As a DL baseline pre-trained BERT \cite{bert} model and Bert tokenizer with a maximum sequence length of 256 inputs have been selected for finetuning. The model has been trained for 10 epochs using Adam optimizer with a learning rate of $10^{-5}$. 
    \item \textit{LNN baseline}. The predicates for the LNN baseline have been selected randomly from Similarity predicates. The number of predicates for each class varies from 340 to 380 predicates.
    \item \textit{Frequency pruning models}. Several models with different frequencies have been trained to examine the effectiveness of the frequency pruning methods. $F>Threshold$ stands for the model with predicates repeating with a frequency higher than the threshold value. The $F>5 $ $balanced$ ensures that classes are balanced in terms of predicates. The remaining predicates have been chosen from a lower frequency.
     \item \textit{Exclusive pruning models}. The exclusive pruning method is used in combination with similarity and frequency pruning methods. In the simulations, Frequency pruning prunes predicates that repeat only once. Then the exclusive pruning removes all the repeating  predicates between classes. 
\end{itemize}

\begin{figure}[h]
    \centering
    \includegraphics[width=0.5\textwidth]{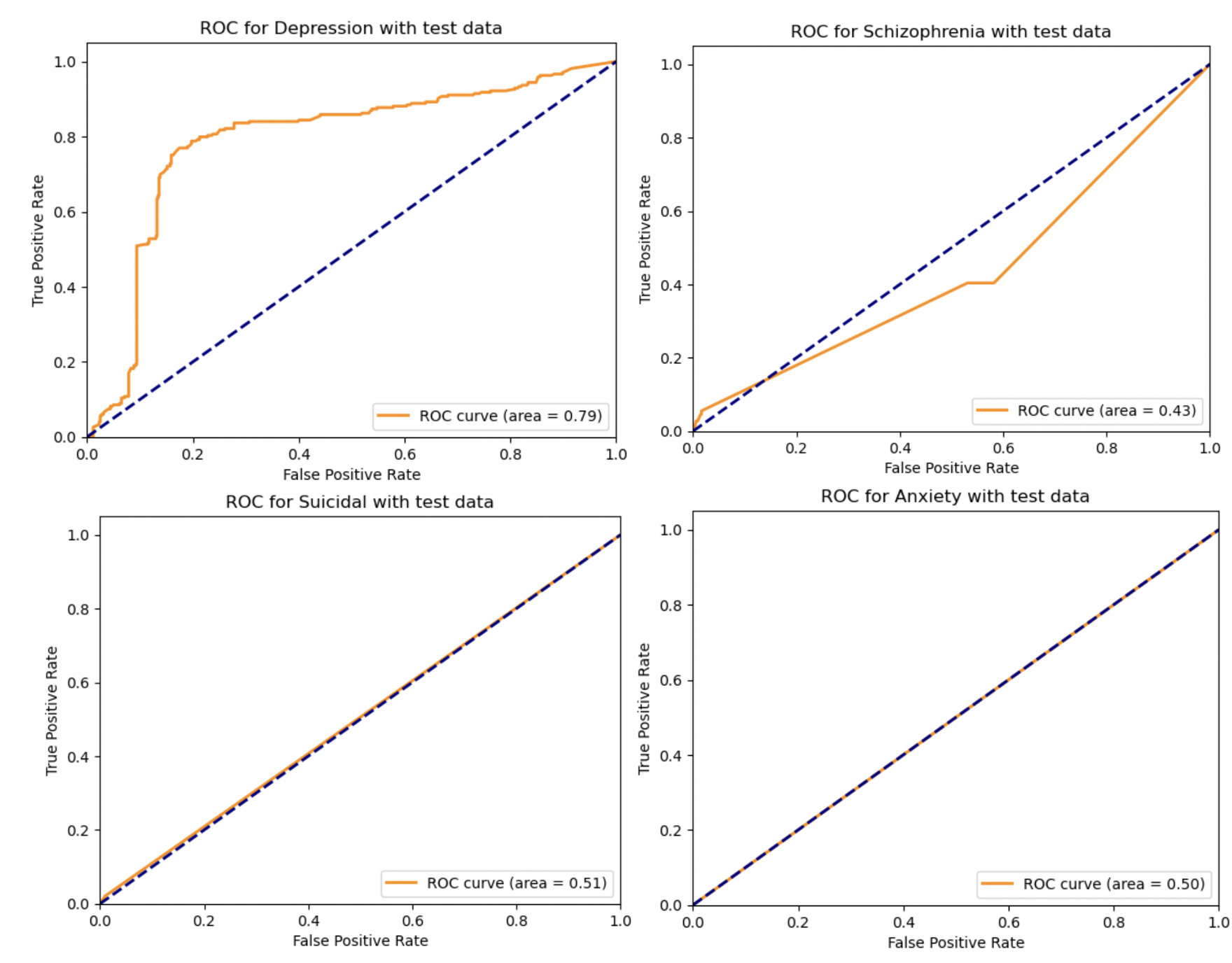}
    \caption[AUC ROC curves for each class in testing.]{AUC ROC curves for each class in testing.}
    \label{fig:LNN}
\end{figure}

According to the Table \ref{table:auc}, Frequency predicates does not have a significant effect on model performance when they are applied alone, since the are under (AUC) the ROC curve is around 0.5, which is close to the random classifier. Moreover, LNN baseline with 1000 predicates and 10000 training samples performed surprisingly well for the Anxiety class, achieving AUC of 0.76.   Baseline DL model has AUC scores higher than 0.72 for all classes when treated as a binary classifier for each class. However, since therapist cannot use this data explicitly, the accuracy for the multi-class classification has higher importance for this case and DL model can provide only 58\% accuracy in such setting. Exclusive predicates model have shown a good performance overall. It  reached AUC of 0.79 for the depression class and 0.57 for schizophrenia. 

\begin{table*}[]
\centering
\begin{tabular}{|l|c|c|cccc|}
\hline
                                                                                                    & \multicolumn{1}{l|}{\textbf{\begin{tabular}[c]{@{}l@{}}\# of training \\ samples\end{tabular}}} & \multicolumn{1}{l|}{\textbf{\begin{tabular}[c]{@{}l@{}}\# of   \\ predicates\end{tabular}}} & \multicolumn{1}{l|}{\textbf{\begin{tabular}[c]{@{}l@{}}Suicidal\\ (AUC)\end{tabular}}} & \multicolumn{1}{l|}{\textbf{\begin{tabular}[c]{@{}l@{}}Depression\\ (AUC)\end{tabular}}} & \multicolumn{1}{l|}{\textbf{\begin{tabular}[c]{@{}l@{}}Anxiety\\ (AUC)\end{tabular}}} & \multicolumn{1}{l|}{\textbf{\begin{tabular}[c]{@{}l@{}}Schizophrenia\\ (AUC)\end{tabular}}} \\ \hline
\textit{BaselineLNN}                                                                                & 10000                                                                                           & 1000                                                                                        & \multicolumn{1}{c|}{0.50}                                                              & \multicolumn{1}{c|}{0.50}                                                                & \multicolumn{1}{c|}{0.76}                                                             & 0.52                                                                                        \\ \hline
\multirow{2}{*}{\textit{BaselineDL}}                                                                & \multirow{2}{*}{10000}                                                                          & \multirow{2}{*}{N/A}                                                                        & \multicolumn{1}{c|}{0.73}                                                              & \multicolumn{1}{c|}{0.83}                                                                & \multicolumn{1}{c|}{0.81}                                                             & 0.72                                                                                        \\ \cline{4-7} 
                                                                                                    &                                                                                                 &                                                                                             & \multicolumn{4}{c|}{Accuracy   for multiclass = 0.58}                                                                                                                                                                                                                                                                                                                   \\ \hline
\textit{F \textgreater 5}                                                                           & 3947                                                                                            & 87                                                                                          & \multicolumn{1}{c|}{0.55}                                                              & \multicolumn{1}{c|}{0.58}                                                                & \multicolumn{1}{c|}{0.55}                                                             & 0.52                                                                                        \\ \hline
\textit{\begin{tabular}[c]{@{}l@{}}F\textgreater{}5.   \\ Balanced\end{tabular}}                 & 3947                                                                                            & 141                                                                                         & \multicolumn{1}{c|}{0.55}                                                              & \multicolumn{1}{c|}{0.59}                                                                & \multicolumn{1}{c|}{0.52}                                                             & 0.55                                                                                        \\ \hline
\textit{F   \textgreater 3}                                                                      & 3605                                                                                            & 349                                                                                         & \multicolumn{1}{c|}{0.54}                                                              & \multicolumn{1}{c|}{0.56}                                                                & \multicolumn{1}{c|}{0.53}                                                             & 0.52                                                                                        \\ \hline
\textit{F\textgreater{}6}                                                                        & 3947                                                                                            & 81                                                                                          & \multicolumn{1}{c|}{0.55}                                                              & \multicolumn{1}{c|}{0.53}                                                                & \multicolumn{1}{c|}{0.56}                                                             & 0.50                                                                                        \\ \hline
\textit{\begin{tabular}[c]{@{}l@{}}Exclusive   predicates  \\ and F\textgreater{}1\end{tabular}} & 3947                                                                                            & 981                                                                                         & \multicolumn{1}{c|}{0.51}                                                              & \multicolumn{1}{c|}{0.79}                                                                & \multicolumn{1}{c|}{0.50}                                                             & 0.43                                                                                        \\ \hline
\end{tabular}

\caption{  AUC ROC scores for DL baseline, LNN baseline and proposed pruning methods .}
\label{table:auc}
\end{table*}


\section{Discussions and Future Work}

Scaling of the LNN is a significant issue which requires selection of the right predicates. Pruning of the predicates essentially limits the knowledge base of the LNN, thus it is important to understand the effect of the predicates on model performance. Frequency predicate models have not shown great results, the possible explanation for that behavior can be found in predicate analysis. The analysis shows that predicates with higher frequencies also tend to be inclusive for several classes. Such predicates might be extracted from common dialogue phrases, that are common to regular conversations.  Thus, it is more difficult to learn for LNN in such circumstances and it might lead to a behavior similar to the random classifiers'.  Moreover, variation in the frequency thresholds did not affect the overall performance of the LNN model. Thus, it can be concluded that frequency predicates cannot provide a quality selection of the predicates when they are applied alone. Furthermore, in the case of exclusive predicates, the model has learned depression class better than others. It can be explained by the depression class possessing more exclusive predicates compared to other classes. Interestingly, the model has learned to identify non-schizophrenia samples better than schizophrenia samples. Possible reasoning for that is fewer predicates for schizophrenia compared to other classes. Furthermore, some mental disorders have the same symptoms, and exclusive pruning eliminate such predicates from the training, which might lead to limited diagnostic abilities. Thus, exclusive predicates should be combined with other methods to provide trade-off between generalization and exclusivity of predicates.

Another challenge of this line of work is the usage of AI for mental disorder diagnosis. As pointed out in \cite{lin2022computational}, one significant challenge is related to the privacy and security of patient data. To train the model, the system requires access to sensitive patient data, which must be protected from unauthorized access or misuse. There is also a concern that the use of AI in mental health diagnosis may lead to the stigmatization of individuals with mental disorders. In this work, we have deidentified all the sessions and all the transcripts are obtained under proper license and consent. We would also like to point out that the system may not work for all individuals, which could lead to misdiagnosis or lack of diagnosis, leading to harm to the patient. Therefore, the ethical challenge lies in ensuring the system's reliability, fairness, and transparency and balancing the use of AI with the need for human involvement in mental health diagnosis and treatment, as part of the future work.

\begin{table*}[h]
\centering
\begin{tabular}{|l|l|l|l|l|}
\hline
                       & \textbf{\begin{tabular}[c]{@{}l@{}}Grouping of \\ predicates\end{tabular}}    & \textbf{\begin{tabular}[c]{@{}l@{}}Top 1\\ weight\end{tabular}} & \textbf{\begin{tabular}[c]{@{}l@{}}Top 2 \\ weight\end{tabular}} & \textbf{\begin{tabular}[c]{@{}l@{}}Top 3\\  weight\end{tabular}}    \\ \hline
\textbf{Depression}    & \begin{tabular}[c]{@{}l@{}}Related to \\ first-person \\ actions\end{tabular} & Do\_i                                                           & \begin{tabular}[c]{@{}l@{}}Come\\ \_they\end{tabular}            & \begin{tabular}[c]{@{}l@{}}Resemble\\ \_what\end{tabular}           \\ \hline
\textbf{Anxiety}       & \begin{tabular}[c]{@{}l@{}}Related to \\ feelings\end{tabular}                & get\_it                                                         & look\_it                                                         & \begin{tabular}[c]{@{}l@{}}have-rel-\\ role\_my \\ mom\end{tabular} \\ \hline
\textbf{Schizophrenia} & \begin{tabular}[c]{@{}l@{}}Related to \\ medical \\ terms\end{tabular}        & give\_me                                                        & \begin{tabular}[c]{@{}l@{}}resemble\\ \_things\end{tabular}      & do\_it                                                              \\ \hline
\textbf{Suicidal}      & \begin{tabular}[c]{@{}l@{}}Related to \\ third-person \\ actions\end{tabular} & \begin{tabular}[c]{@{}l@{}}have-manner\\ \_sense\end{tabular}   & put\_it                                                          & \begin{tabular}[c]{@{}l@{}}do\_\\ everything\end{tabular}           \\ \hline
\end{tabular}
\caption{Analysis of the semantics of the predicates for the each class.}
\label{table:expl}
\end{table*}

The main advantage of the LNN over DL is in its explainability. It is possible to extract predicates with high weights for the each class and to examine which predicates contribute to the result significantly. Table \ref{table:expl} shows the predicate semantics analysis for each class after the training. Predicates of depression and anxiety suicidal classes are mostly related to the first-person and third-person actions respectively, while people with anxiety tend to talk about feelings more. In addition, predicates of the schizophrenia class tend to relate to the medical terms. This overlaps with overall content of the transcripts and predicates that posses high weights can be used to give insights to therapist during the diagnosis of the patients. 

\subsection{Future work}
Overall, it is evident that predicates are too specific from the number of predicates with a frequency of 1, which might be a possible explanation for the poor performance of the model overall. Therefore, they might require some generalization of the predicates. One of the promising methods for that is to use synonym-based predicates. By using thesaurus dictionaries, it is possible to cluster all the keys and values of the AMR representations and use only one variants for the synonyms. That way, it might be possible to reduce the number of predicates significantly and achieve their generalization. 

Another possible way to enhance the model is the explore LNN and DL hybrid approach. By using LNN scores it is possible to train some dense layers with SoftMax to predict classes in the multiclass setting. In a such way it will be more convenient to compare LNN results with DL solutions while keeping the explainability of the LNN.

\section{Conclusion}
Mental disorders are a significant issue that is affecting more people every year. Therefore, explainable AI mental disorder diagnosis through utterance classification can aid the therapist in their practice. In this work, a supervised learning setting for the LNN has been proposed to address this issue. Moreover, predicate pruning methods based on the similarity, frequency, and exclusivity of the predicates are analyzed in terms of training performance. Overall, the model trained with exclusive predicates shows the best results among the pruning methods, and acheived AUC ROC of 0.79 for the depression disorder.   Finally, explainability of the LNN diagnosis has been shown by analyzing significant predicates for each class and extracting the predicates with high weights.

\bibliography{custom}
\bibliographystyle{acl_natbib}



\end{document}